\documentclass[letterpaper]{article} 
\usepackage{aaai2026}  
\usepackage{times}  
\usepackage{helvet}  
\usepackage{courier}  
\usepackage[hyphens]{url}  
\usepackage{graphicx} 
\usepackage{natbib}  
\usepackage{caption} 
\usepackage{algorithm}
\usepackage{algorithmic}
\usepackage{amsmath}
\usepackage{dsfont}
\usepackage{booktabs}   
\usepackage{amsfonts}

\usepackage{newfloat}
\usepackage{listings}
\DeclareCaptionStyle{ruled}{labelfont=normalfont,labelsep=colon,strut=off} 
\floatstyle{ruled}
\newfloat{listing}{tb}{lst}{}
\floatname{listing}{Listing}
\title{Causal Reinforcement Learning based Agent-Patient
Interaction \\ with Clinical Domain Knowledge}
\author{
    Wenzheng Zhao\textsuperscript{\rm 1},
    Ran Zhang\textsuperscript{\rm 2},
    Ruth Palan Lopez\textsuperscript{\rm 3},
    Shu-Fen Wung\textsuperscript{\rm 4},
    Fengpei Yuan\textsuperscript{\rm 1}
}
\affiliations{
  \textsuperscript{\rm 1}Dept. of Robotics Engineering, Worcester Polytechnic Institute\\
  \textsuperscript{\rm 2}Dept. of Electrical and Computer Engineering, University of North Carolina at Charlotte\\
  \textsuperscript{\rm 3}School of Nursing, MGH Institute of Health Professions\\
  \textsuperscript{\rm 4}School of Nursing, University of California, Davis\\
  wzhao8@wpi.edu, rzhang8@charlotte.edu, rlopez@mghihp.edu, swung@health.ucdavis.edu , fyuan3@wpi.edu
}

\begin{document}

\maketitle

\begin{abstract}
Reinforcement Learning (RL) faces significant challenges in adaptive healthcare interventions, such as dementia care, where data is scarce, decisions require interpretability, and underlying patient-state dynamic are complex and causal in nature. In this work, we present a novel framework called Causal structure-aware Reinforcement Learning (CRL) that explicitly integrates causal discovery and reasoning into policy optimization. This method enables an agent to learn and exploit a directed acyclic graph (DAG) that describes the causal dependencies between human behavioral states and robot actions, facilitating more efficient, interpretable, and robust decision-making.
We validate our approach in a simulated robot-assisted cognitive care scenario, where the agent interacts with a virtual patient exhibiting dynamic emotional, cognitive, and engagement states. The experimental results show that CRL agents outperform conventional model-free RL baselines by achieving higher cumulative rewards, maintaining desirable patient states more consistently, and exhibiting interpretable, clinically-aligned behavior. We further demonstrate that CRL’s performance advantage remains robust across different weighting strategies and hyperparameter settings. In addition, we demonstrate a lightweight LLM-based deployment: a fixed policy is embedded into a system prompt that maps inferred states to actions, producing consistent, supportive dialogue without LLM finetuning. Our work illustrates the promise of causal reinforcement learning for human-robot interaction applications, where interpretability, adaptiveness, and data efficiency are paramount.
\end{abstract}

\begin{links}
  \link{Code}{Link-will-be-released-upon-publication}
  \link{Datasets}{Link-will-be-released-upon-publication}
\end{links}
  
\section{Introduction}

\begin{figure}[!htp]
  \centering
  \includegraphics[width=\linewidth]{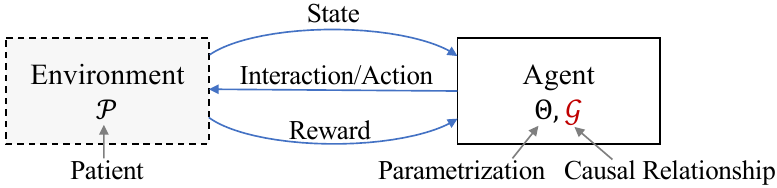}  
  \caption{The Agent-Environment interaction from Causal Reinforcement Learning (CRL).}
  \label{fig:crl-frame}
\end{figure}

Personalized and adaptive AI is critical for effective human-agent interaction in healthcare, particularly in sensitive domains like dementia care where patient needs vary significantly and evolve over time. While reinforcement learning (RL) offers a powerful framework for sequential decision-making, its application in healthcare faces fundamental challenges: extensive data requirements, poor sample efficiency, and lack of interpretability—limitations that become prohibitive in clinical contexts where data is scarce, exploration must be safe, and decisions require clinical justification.

Causal reasoning has emerged as a promising approach to address these limitations in RL~\cite{pearl2009causality,seitzer2021causal}.
By enabling agents to reason about interventions and counterfactual outcomes, causal methods can significantly improve both sample efficiency and interpretability.
However, despite this potential, the integration of causal reasoning into human-agent interaction systems for clinical environments remains underexplored. 
Current approaches in dementia care exemplify this gap: existing system rely either on model-free RL methods that lack transparency and clinical interpretability~\cite{yuan2021learning}, or on rigid, rule-based or data-heavy agents that cannot adapt to individual patient needs~\cite{hung2019benefits,bevilacqua2023social}. 
This dichotomy between data-driven adaptability and safe, interpretable decision-making presents a critical barrier to deploying AI in real-world clinical settings.

We address this gap by introducing a Causal RL (CRL) framework that integrates structured clinical domain knowledge into the agent's learning loop. Our approach leverages causal discovery to model latent cognitive-emotional states and action effects, constructing a causal world model that explicitly guides policy learning through causal inference. This enables the agent to reason counterfactually about action outcomes, leading to more data-efficient learning and clinically-justifiable decisions. Our main contributions are:

\begin{itemize}
    \item We propose a novel causality-aware RL framework that systematically embeds clinical domain knowledge through causal discovery and reasoning. To the best of our knowledge, this is the first work to integrate expertise in this manner for efficient reasoning in agent-patient interactions.
    
    \item We demonstrate that our causality-informed RL agent achieves improved sample efficiency and generalization by inferring the consequences of actions, thereby reducing its dependence on exhaustive and potentially unsafe trial-and-error.
    
    \item We validate our approach in a simulated patient-robot interaction environment for cognitive rehab therapy (reminiscence therapy). Experimental results show that causal-informed agent significantly outperforms traditional RL baselines in both performance, sample efficiency, and transparency.
\end{itemize}

\section{Related Work}

\subsection{Causal Reinforcement Learning}

Integrating causal reasoning into reinforcement learning (RL) has shown significant potential in improving sample efficiency, policy robustness, and interpretability~\cite{pearl2009causality, deng2023causal}.Recent efforts have incorporated causal reasoning into reinforcement learning to improve sample efficiency and generalization~\cite{seitzer2021causal,wang2021provably}. These methods often rely on discovering or leveraging causal graphs to guide exploration and avoid spurious correlations. However, these causal enhanced RL framework have yet to be thoroughly explored within interactive human-centered domains such as cognitive care scenarios. Our work fills this critical gap by integrating causal discovery with RL in a realistic dementia care setting, demonstrating enhanced policy performance and interpretability. 

\subsection{Causal Reasoning and RL in Human-Robot Interaction (HRI)}
RL has been applied in HRI to learn personalized dialogue strategies or adaptive interaction behaviors, especially for socially assistive robotics~\cite{yuan2021learning}. However, most existing approaches are model-free and lack causal interpretability. In contrast, several recent works have begun to explore the integration of causal reasoning mechanisms into HRI systems: ~\cite{spitale2025exploring} proposed an approach for robotic mental well-being coaching that employs structural equation models to characterize causal relationships between robot behaviors, facial valence and speech duration. Their framework enables the robot to adapt its strategies based on inferred causal structures, improving interaction outcomes. ~\cite{tung2024causal} proposed a method based on mutual information estimation to quantify the causal influence of robot actions on human behavioral responses, underscoring the significance of identifying when and how robot behavior causally shapes human interaction dynamics for more effective HRI strategy design. These studies demonstrate a growing trend in HRI research toward embedding causal reasoning to improve interpretability, adaptiveness, and robustness. Moreover, their methodologies provide valuable ideas and motivation for the causal reinforcement learning framework proposed in this work.

\subsection{Modeling Cognitive States in Assistive Robotics}

Robotic systems designed to assist individuals with cognitive impairments frequently model user cognitive and emotional states to enable personalized interaction strategies~\cite{chung2008carers}.~\cite{yuan2025integrating} proposed a framework that integrates reinforcement learning (RL), large language model (LLM)-based user simulators, and clinical domain knowledge to represent the evolving cognitive and emotional trajectories of persons living with dementia (PLWDs), allowing socially assistive robots to dynamically tailor their conversational and assistance behaviors in simulated care scenarios. Earlier, ~\cite{rudzicz2015speech} conducted real-world speech interaction experiments with older adults with Alzheimer’s disease, demonstrating that frequent confusion behaviors, such as ignoring prompts, are prevalent during human-robot interaction. These findings collectively underscore the importance of accurately modeling cognitive state fluctuations to inform and optimize robot behavior in assistive contexts.

\subsection{Behavioral-Level Human-AI Interaction in Dementia Care}

Social robots have been increasingly explored in dementia care as tools for emotional and behavioral support. For example, robot-assisted therapy has been reported to improve mood, encourage social interaction, and facilitate communication among individuals with dementia~\cite{bevilacqua2023social, hung2019benefits}. Beyond physical robot~\cite{rashid2023effectiveness}, conversational agents have also been investigated as tools to deliver reminiscence therapy or cognitive training to persons living with dementia (PLWDs). More recently,~\cite{chorbadzhieva2025exploring} introduced an AI-driven reminiscence companion powered by large language models to adaptively engage PLWDs in long-term care environments. However, these studies highlight the growing interest in behavioral-level human-AI interaction for dementia care. However, existing approaches remain limited by heavy reliance on scripted or data-intensive methods, and rarely incorporate causal or reinforcement learning frameworks. Our work addresses this gap by introducing a causal reinforcement learning paradigm that enables adaptive, interpretable, and sample-efficient strategies tailored to the evolving cognitive and emotional states of PLWDs.

\section{Proposed Causal Reinforcement Learning Framework}

In our design, as shown in Algorithm~\ref{alg:crl}, we introduce causal reasoning into the RL loop (Q-learning-based) to 
reduce the amount of data required for training and to improve policy generalization. By explicitly modeling the causal relationships within the environment, the agent can leverage clinical domain knowledge to make informed decisions, rather than relying solely on extensive trial-and-error exploration. This capability is particularly important in high-risk, high-cost domains such as healthcare, where safe, efficient, and adaptive online learning is essential. We implemented and evaluated our Algorithm~\ref{alg:crl} using the simulation model of person with dementia (PwD)-robot interaction in the context of reminiscence therapy~\cite{yuan2021learning}.

\begin{algorithm}[t]
\caption{Causal Structure-Aware Q-learning (CRL)}
\label{alg:crl}
\small
\begin{algorithmic}[1]
\STATE Initialize $Q(s,a) \gets 0$, for all $s \in S$, $a \in A$
\FOR{each epoch $e=1$ to $E$}
    \FOR{each episode $i=1$ to $N$}
        \STATE Initialize state $s_0 = [NR, Neu, No]$, $done \gets \text{False}$
        \WHILE{not $done$}
            \IF{agent observed negative emotion or confusion persistence}
                \STATE $a_t \gets a_6$ (GiveChoice)
            \ELSE
                \STATE Compute mixed policy weights $(w_{RL}, w_{DAG})$ based on CRL-static/dynamic rule
                \STATE Sample coin $\in [0,1)$
                \IF{coin $\leq w_{RL}$}
                    \STATE $a_t \gets$ action from $\epsilon$-greedy w.r.t. $Q$
                \ELSIF{coin $\leq w_{RL} + w_{DAG}$}
                    \STATE $a_t \gets$ action suggested by learned DAG
                \ELSE
                    \STATE $a_t \gets$ random action
                \ENDIF
            \ENDIF
            \STATE Execute $a_t$, observe reward $r_t$, next state $s_{t+1}$
            \STATE Update $Q(s_t,a_t)$
            \STATE $s_t \gets s_{t+1}$
        \ENDWHILE
    \ENDFOR
\ENDFOR
\end{algorithmic}
\end{algorithm}

\subsection{State and Action Space Definition}
\label{state_action_space}
Based on the simulation environment of PwD-robot interaction during reminiscence therapy, the Markov Decision Process (MDP) is defined over a structured state space that captures three key dimensions of the PwD's condition:
\textbf{(i) Response Relevance}: the relevance of the PwD’s response to the robot’s prompt, categorized as no response (NR), irrelevant response (IR), or relevant response (RR), i.e., $\{0: \texttt{NR}, 1: \texttt{IR}, 2: \texttt{RR}\}$;\textbf{(ii) Emotional State (Pleasure)}: the detected emotion level, categorized as negative, neutral, or positive, i.e., $\{-1: \texttt{Neg}, 0: \texttt{Neu}, 1: \texttt{Pos}\}$;
\textbf{(iii) Confusion Condition}: whether the PwD is confused, with two possible values, $\{0: \texttt{No}, 1: \texttt{Yes}\}$. The complete state space \( S \) is the Cartesian product of the three dimensions above, resulting in a total of \( 3 \times 3 \times 2 = 18 \) discrete states. The action space $A$ includes six potential robot actions: 
\{0: \texttt{EasyPrompt}, 
1: \texttt{ModeratePrompt}, 
2: \texttt{DifficultPrompt}, 
3: \texttt{Repeat}, 
4: \texttt{Explain}, 
5: \texttt{Comfort}\}. 
These actions are designed to either stimulate engagement (e.g., by prompting the PwD with varying difficulty), support comprehension (e.g., repeating or explaining prompts), or provide emotional reassurance.

\subsection{Reward Function Design}
\label{reward_function}

The reward function \( R(s_t,a_t,s_{t+1}) \) is designed to align with the dual objectives of robot-assisted reminiscence therapy (RT): (i) encouraging verbal engagement from the person with dementia (PwD), and (ii) maintaining the PwD in a generally positive affective and cognitive state. To achieve these objectives, the reward function explicitly accounts for three dimensions of the PwD's status (See Section~\ref{state_action_space}). Let $s_t=(RP_t,E_t,C_t)$, $a_t\in A$, and the subsequent state $s_{t+1}=(RP_{t+1},E_{t+1},C_{t+1})$. The per–time-step reward is

\begin{align}
r_t &= r_{\text{resp}}(RP_{t+1}, a_t)
      + r_{\text{emotion}}(E_{t+1})
      + r_{\text{conf}}(C_{t+1}) \nonumber\\
    &\quad + \eta
      + \delta\,\mathds{1}\{\text{NewTrigger}_t\} \nonumber\\
    &\quad + \lambda_{\text{stop}}\,\mathds{1}\{\text{EarlyStop}_t\}
      + \lambda_{\text{goal}}\,\mathds{1}\{\text{GoalReached}_t\}.
\label{eq:reward_sum}
\end{align}

Here $\eta$ is an optional per-round shaping term (default $\eta=0$), 
$\delta$ is the bonus for using a new memory trigger, 
$\lambda_{\text{stop}}=-200$ is the early-termination penalty, and 
$\lambda_{\text{goal}}=15$ is the bonus for completing all triggers or reaching the $50$-round cap. 

\textbf{Response–action term.}
Let $\mathcal{O}=\{a_0,a_3,a_4,a_5,a_6\}$ denote “other” actions, 
and $a_1$ (ModeratePrompt), $a_2$ (DifficultPrompt). Then

\begin{align}
r_{\text{resp}}(RP_t,a_t) &=
\begin{cases}
-2, & RP_t=\text{NR},\; \forall a_t\\[2pt]
\;0.75, & RP_t=\text{IR},\; a_t=a_1\\
\;1.75, & RP_t=\text{IR},\; a_t=a_2\\
\;0.30, & RP_t=\text{IR},\; a_t\in\mathcal{O}\\[2pt]
\;2, & RP_t=\text{RR},\; a_t=a_1\\
\;3, & RP_t=\text{RR},\; a_t=a_2\\
\;0.75, & RP_t=\text{RR},\; a_t\in\mathcal{O}
\end{cases}.
\label{eq:resp_action}
\end{align}

\textbf{Emotion and confusion terms.}
\begin{align}
r_{\text{emotion}}(E_t) &=
\begin{cases}
-3, & E_t = \text{Neg} \\
 1, & E_t = \text{Neu} \\
 2, & E_t = \text{Pos}
\end{cases} \label{eq:emotion}\\[6pt]
r_{\text{conf}}(C_t) &=
\begin{cases}
-2.5, & C_t = \text{Yes} \\
  2,   & C_t = \text{No}
\end{cases}
\label{eq:conf}
\end{align}

\subsection{Causal Relationship Learning Algorithm Design}
\label{Causal Relationship Learning Algorithm Design}

To support causal-aware policy learning, we first generate simulated interaction trajectories where patient states $(RP, E, C)$ evolve in response to random robot actions $A_t$. This data enables causal discovery (e.g., PC, LiNGAM~\cite{JMLR:v25:22-1258,dowhy}) to infer a directed acyclic graph (DAG) linking current states, actions, and subsequent states. 
\textit{See Appendix~\ref{appendix:causal_details} for full dataset construction details.}
We then estimate Conditional Average Treatment Effects (CATE) using \texttt{EconML}~\cite{econml}, quantifying, given different patient states, the relative impact of each action compared to a baseline action, $a_0$ (EasyPrompt), on reward, as fomulated in Equation~\ref{eq:causalInference}. The learned causal relationships provide fine-grained insights into how actions influence engagement and emotion across different patient states and serves as clinical domain knowledge to guide RL policy learning.

\begin{equation}\label{eq:causalInference}
\text{CATE}_{a_i, S} = \mathbb{E}[Y \mid do(A=a_{i, i\neq0}), S] - \mathbb{E}[Y \mid do(A=a_0), S].
\end{equation}

To ensure clinical validity, we also conducted an expert evaluation where Dementia Nurse Practitioners assessed the plausibility of discovered causal patterns (e.g., the effectiveness of \texttt{Comfort} or \texttt{Explain} under confusion). Expert evaluation and feedback from two dementia nurse practitioners confirmed the DAG's alignment with dementia care domain knowledge and suggested directions for refining the state representation, though such refinement is beyond this work's focus.

\subsection{Experiments}

We evaluate CRL in a simulated PwD-robot interaction environment with an 18-state space and 7 robot actions, where the 7th action is deterministic. 

\subsection{Training Details}
All methods were trained for 1500 epochs, each consisting of 30 episodes. 
Evaluation results were averaged over 100 independent test runs using the final policy from the last training epoch. 
This ensures robustness against stochastic variation. 
All methods share identical hyperparameters: learning rate of 0.05, discount factor $\gamma=0.95$, and $\epsilon$-greedy exploration with $\epsilon=0.1$.

\subsection{Baselines and Ablation}
We compared our approach against two representative baselines: \textbf{(i) Model-Free RL}: standard Q-learning agent without access to causal information. \textbf{(ii) DAG-Only}: actions selected purely based on the learned DAG structure, ignoring reward signals. \textbf{(iii) CRL-Static}: fixed weighting of causal DAG guidance and RL signals throughout training. \textbf{(iv) CRL-Dynamic}: adaptive weighting scheme that shifts reliance from DAG to RL over training epochs. Table~\ref{tab:rl_dag_weights} summarizes the RL and DAG weight configurations used in these methods, including how CRL-Dynamic adapts weights over training epochs.
We compare against Model-Free RL, DAG-Only, CRL-Static, and CRL-Dynamic (adaptive weighting). 
The corresponding RL and DAG weight configurations are summarized in Table~\ref{tab:rl_dag_weights}.

\begin{table}[!htp]
\caption{Weights for RL and DAG Components Across Different Methods}
\label{tab:rl_dag_weights}
\setlength{\tabcolsep}{3pt}
\centering
\scriptsize 
\renewcommand{\arraystretch}{0.9}
\begin{tabular}{lccc}
\toprule
\textbf{Method} & \textbf{Weight\_RL $w_{RL}$} & \textbf{Weight\_DAG $w_{DAG}$} & \textbf{Exploration} \\
\midrule
RL & 0.9 & -- & 0.1 \\
CRL\_STATIC & 0.45 & 0.45 & 0.1 \\
CRL\_Dynamic & 0.2 $\rightarrow$ 0.7 & 0.7 $\rightarrow$ 0.2 & 0.1 \\
DAG & -- & 0.9 & 0.1 \\
\bottomrule
\end{tabular}
\end{table}

\section{Results and Discussion}
\label{Experiments_and_Results}

\paragraph{Causal Relationship.}

\begin{figure}[!htp]
  \centering
  \includegraphics[width=0.9\linewidth]{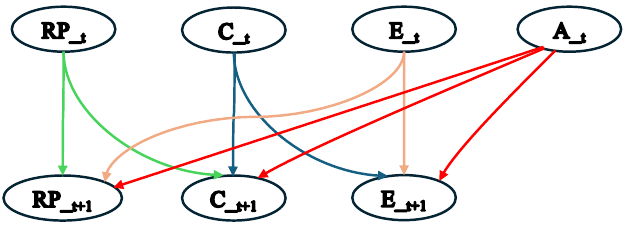}  
  \caption{Learned causal structure illustrating dependencies among robot actions ($A_t$), 
patient’s current states ($S_t=\{RP_t, E_t, C_t\}$), and subsequent states ($S_{t+1}=\{RP_{t+1}, E_{t+1}, C_{t+1}\}$). }
  \label{fig:CATE}
\end{figure}

Figure~\ref{fig:CATE} shows the DAG learned via LiNGAM, revealing dependencies between patient's current states, robot actions, and patient's subsequent states. Expert evaluation confirmed the DAG’s alignment with dementia care knowledge. To illustrate, estimated CATE values indicate that \texttt{Explain} increases expected reward by $+4.3$ under \texttt{IR\_Neu\_Yes} states, while \texttt{Comfort} improves outcomes by $+2.3$ under \texttt{NR\_Neg\_Yes}. These results highlight how supportive actions can modulate cognitive and emotional trajectories, consistent with clinical intuition. \textit{Full CATE tables are provided in Appendix~\ref{appendix:causal_details}}.

\begin{figure}[!htp]
  \centering  \includegraphics[width=0.99\linewidth]   {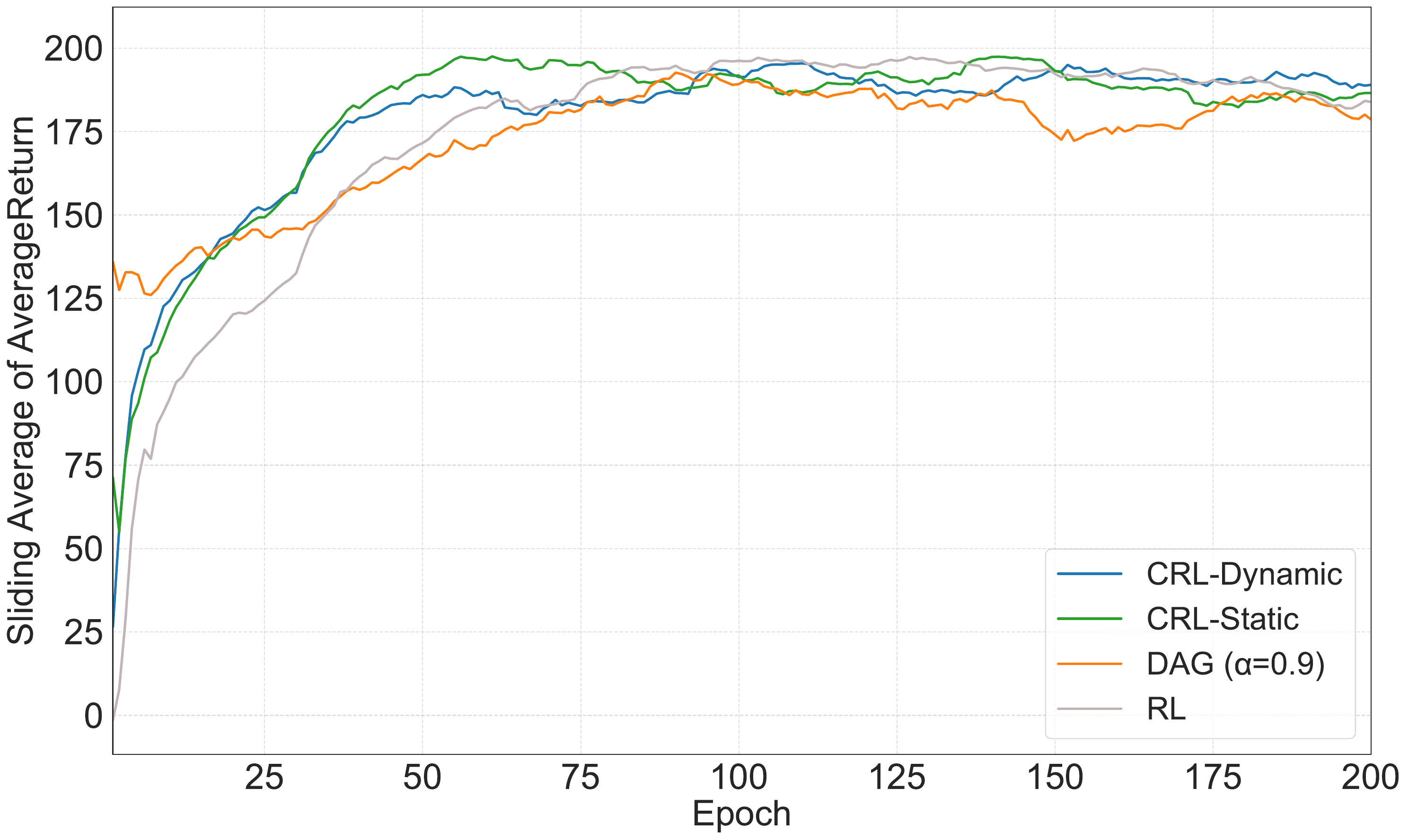}  
  \caption{Smoothed average return across epochs under the final-episode policy of each epoch for different methods (evaluation using RL-only execution).}
  \label{fig:AverageReturn_Comparison_first_200}
\end{figure}

\paragraph{Clinically-Grounded Causal Reasoning Accelerates Policy Learning.}
To disentangle the role of causal priors in accelerating training versus guiding execution, we evaluate policies under two modes:

(i) \textbf{RL-only execution}, where policies learned during training are executed solely based on the Q-learning agent; and 
(ii) \textbf{strategy-consistent evaluation}, where the same weighting scheme used during training (e.g., CRL-Static) is also applied at test time. 
In both cases, CRL-Dynamic achieves superior early performance by leveraging causal priors, while Model-Free RL eventually approaches comparable returns with sufficient training. 
DAG-Only remains limited due to the absence of reward-driven adaptation.

Figure~\ref{fig:AverageReturn_Comparison_first_200} shows the smoothed average return (\textit{Sliding window = 30}) during evaluation. CRL-Dynamic achieves superior early performance, leveraging causal priors to accelerate learning, while Model-Free RL eventually approaches comparable returns with sufficient training. DAG-Only remains limited due to the absence of reward-driven adaptation. \textit{For more comprehensive analysis of the results, please see Appendix~\ref{appendix:results_analysis}.}

\begin{figure*}[!htp]
  \centering
  \includegraphics[width=0.9\linewidth]{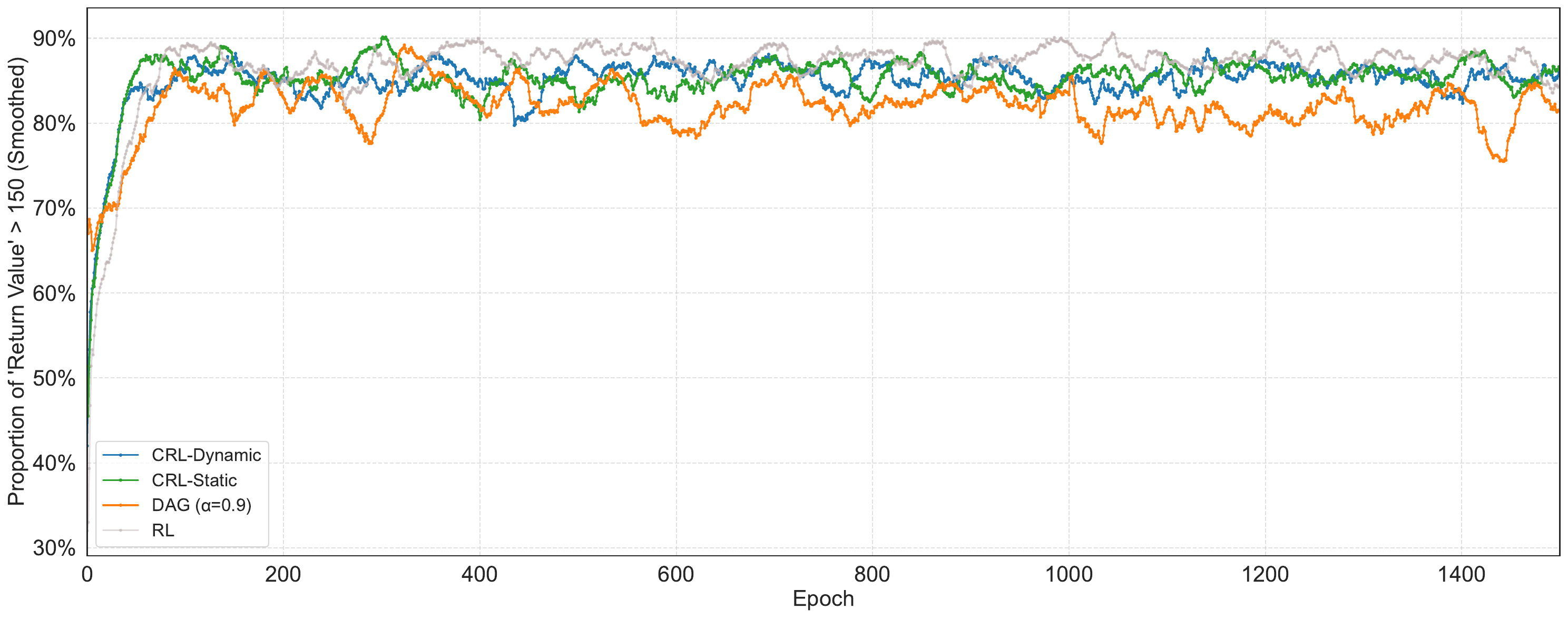}  
  \caption{Proportion of high-return episodes (return $>$ 150) under RL-only execution.}
  \label{fig:more_than_150_rlonly}
\end{figure*}

\begin{figure*}[!htbp]
  \centering  \includegraphics[width=0.87\linewidth]{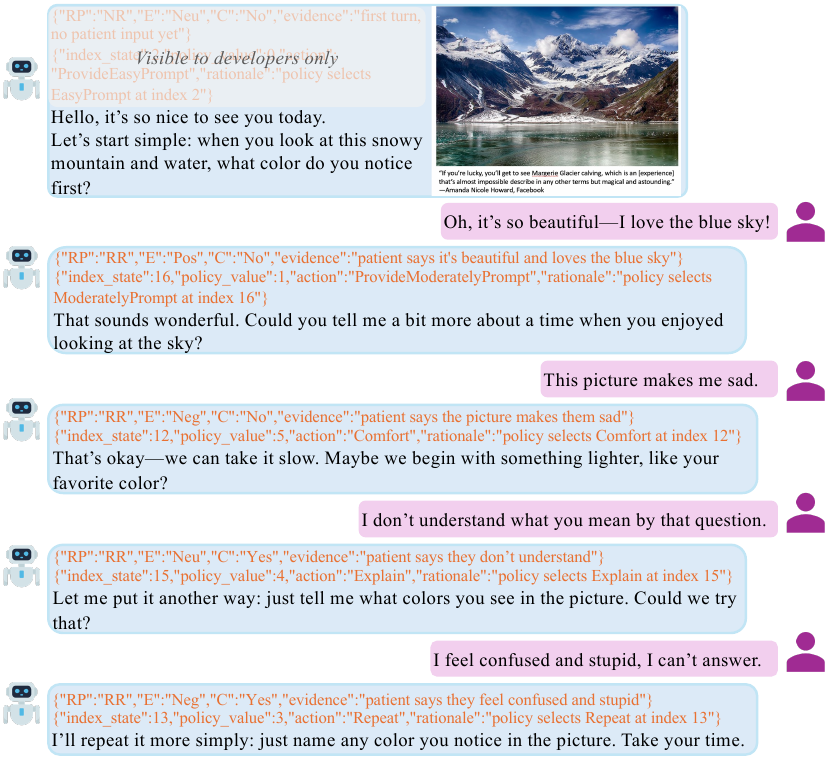}  
  \caption{Example conversation showing CRL policy integration with LLM for patient interaction. State transitions and policy decisions (highlighted in orange) guide the LLM to generate contextually appropriate, emotionally supportive responses.}
  \label{fig:demo}
\end{figure*}

\begin{figure*}[!htp]
  \centering
  \includegraphics[width=0.9\linewidth]{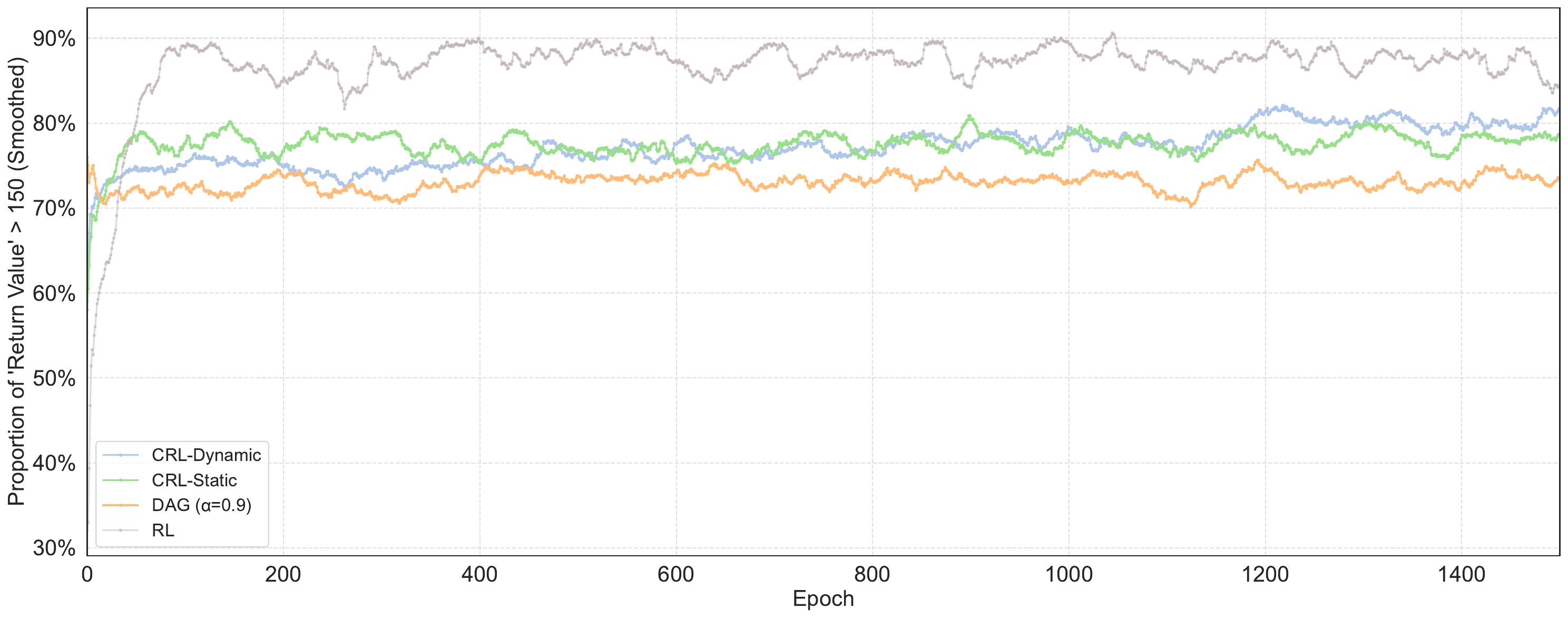}
  \caption{Proportion of high-return episodes (return $>$ 150) under strategy-consistent execution.}
  \label{fig:more_than_150}
\end{figure*}

To complement average-return analysis, we also report the proportion of high-return episodes (threshold $>$ 150). As shown in Figure~\ref{fig:more_than_150_rlonly}: Under RL-only execution, CRL-Dynamic consistently maintains a higher proportion of high-return episodes (return $>$ 150) compared to DAG-only, and approaches the performance of pure RL while converging faster. This indicates that causal priors help accelerate policy learning without compromising asymptotic performance. However, under strategy-consistent evaluation (Figure~\ref{fig:more_than_150}): When policies retain DAG guidance at test time, the proportion of high-return episodes drops significantly for CRL and DAG, whereas RL sustains the highest success rate. This reinforces the insight that causal priors are most beneficial during learning, but should be gradually phased out for execution.

\paragraph{Clinically-Aligned Dialogue Generation via LLM-CRL Integration.}

Figure~\ref{fig:demo} demonstrates the GPT-5 policy’s ability to generate consistent, state-aware dialogue under the CRL-guided framework. The dialogue trace is grounded in the Glacier Bay image shown in Figure~\ref{fig:demo}, which serves as the session hook.
Despite the absence of gradient-based optimization, the fixed policy embedded in the system prompt effectively maps the inferred patient states $(RP, E, C)$ to corresponding supportive actions. 
The model produced coherent and emotionally appropriate responses across five of the six reachable actions in the policy array (\textit{Comfort}, \textit{Explain}, \textit{ProvideModeratelyPrompt}, \textit{Repeat}, \textit{ProvideEasyPrompt}). 
Qualitative inspection shows that GPT-5 sometimes interprets the patient’s inability to elaborate on a prior topic (e.g., struggling to describe the \textit{sky-color}) as a confused state rather than a disengaged or irrelevant one, revealing a \textbf{semantic continuity preference} that prioritizes maintaining conversational flow.
This property enhances interactional smoothness but slightly reduces state discriminability, which aligns with human-like tendencies in empathetic dialogue. 
Overall, the results confirm that the LLM component can reliably enact causal-policy mappings while preserving naturalistic affective behavior, supporting its integration into hybrid CRL–LLM control architectures.
\textit{For more comprehensive analysis of the results, please see Appendix~\ref{appendix:results_analysis}.}

\paragraph{Dialogue Length and Episode Persistence.}

\begin{figure*}[!htbp]
  \centering
  \includegraphics[width=0.9\linewidth]{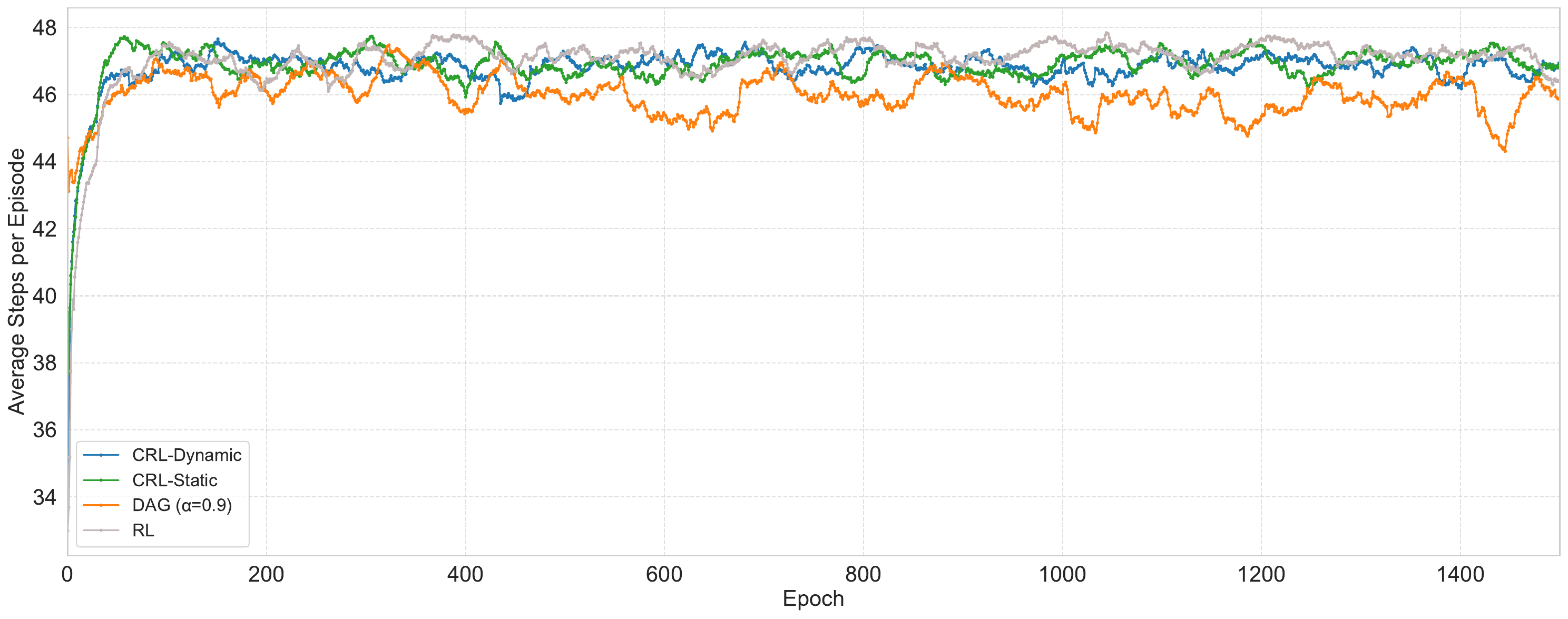}  
  \caption{Average episode length during evaluation under RL-only execution.}
  \label{fig:avg_steps_rl}
\end{figure*}

\begin{figure*}[!htbp]
  \centering
  \includegraphics[width=0.9\linewidth]{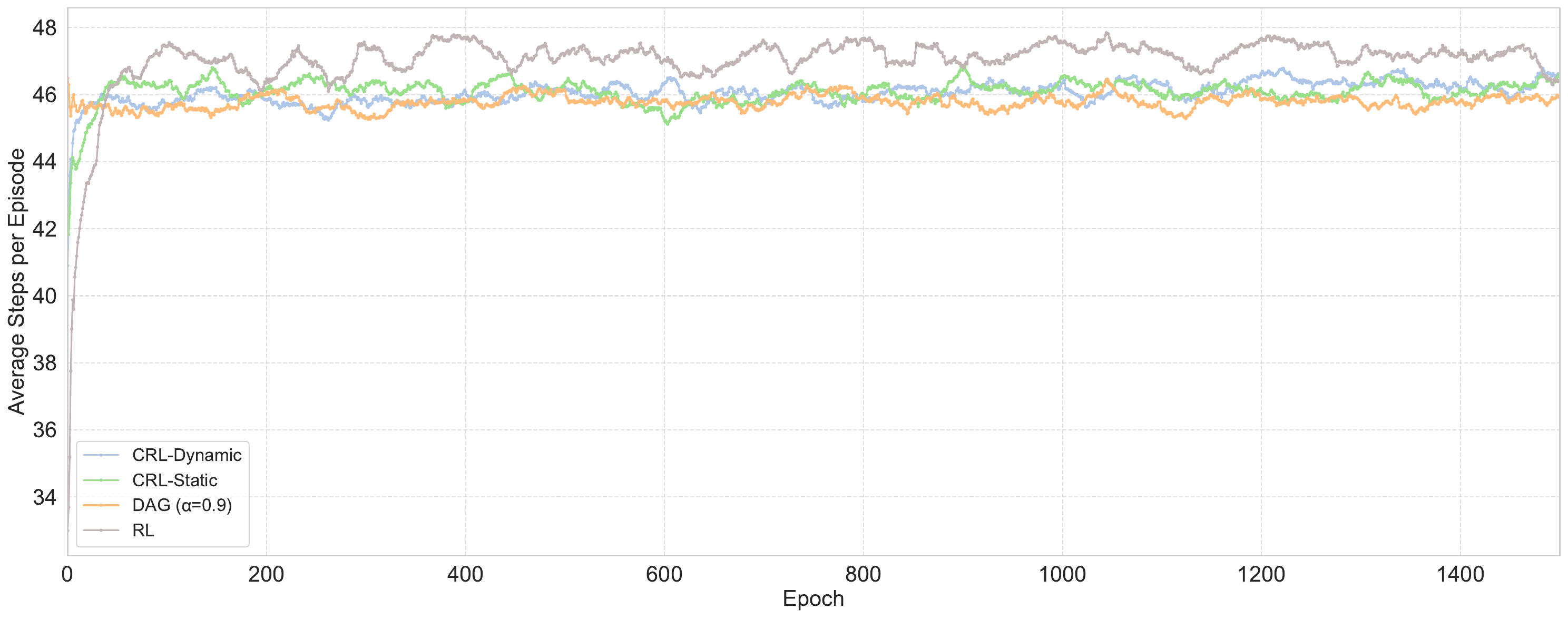}  
  \caption{Average episode length during evaluation under strategy-consistent execution.}
  \label{fig:avg_steps_consistent}
\end{figure*}

\begin{figure*}[!htbp]
  \centering
  \includegraphics[width=0.9\linewidth]{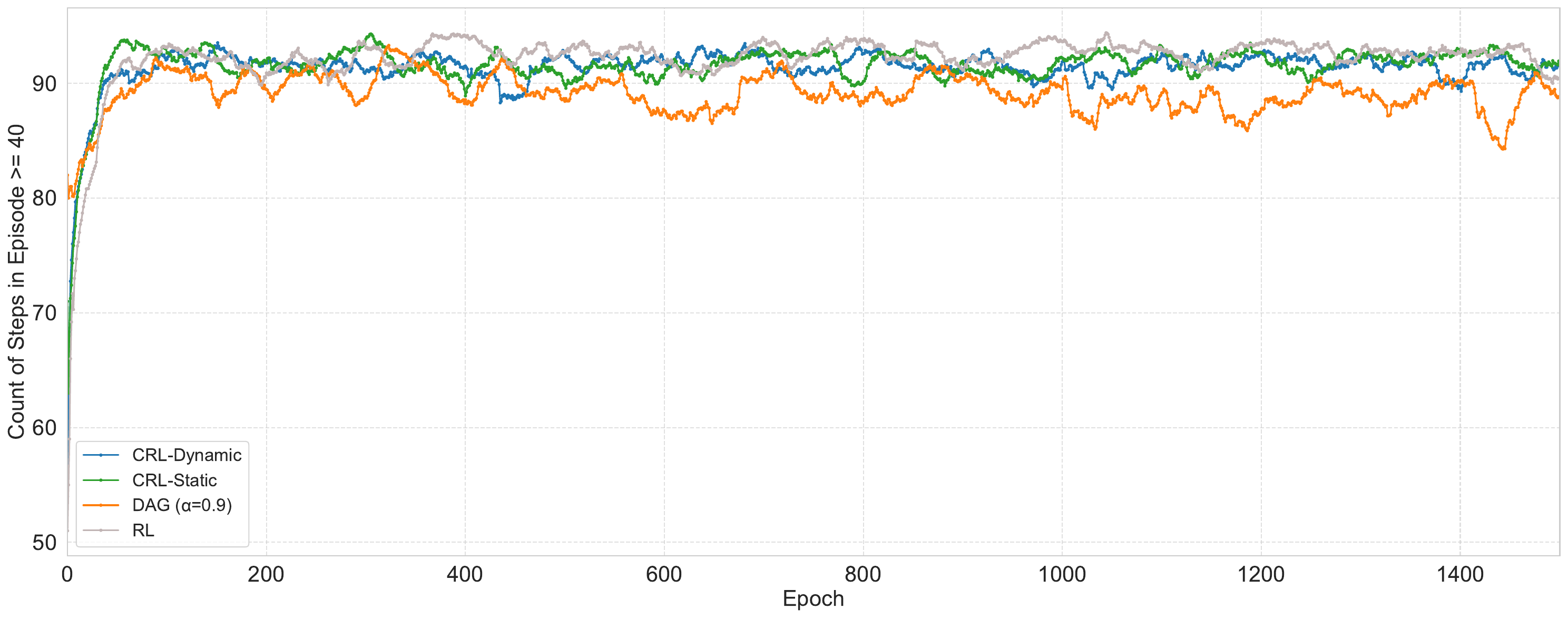}  
  \caption{Proportion of long episodes ($\geq 40$ steps) with RL-only execution.}
  \label{fig:steps_over40_rl}
\end{figure*}

\begin{figure*}[!htbp]
  \centering
  \includegraphics[width=0.9\linewidth]{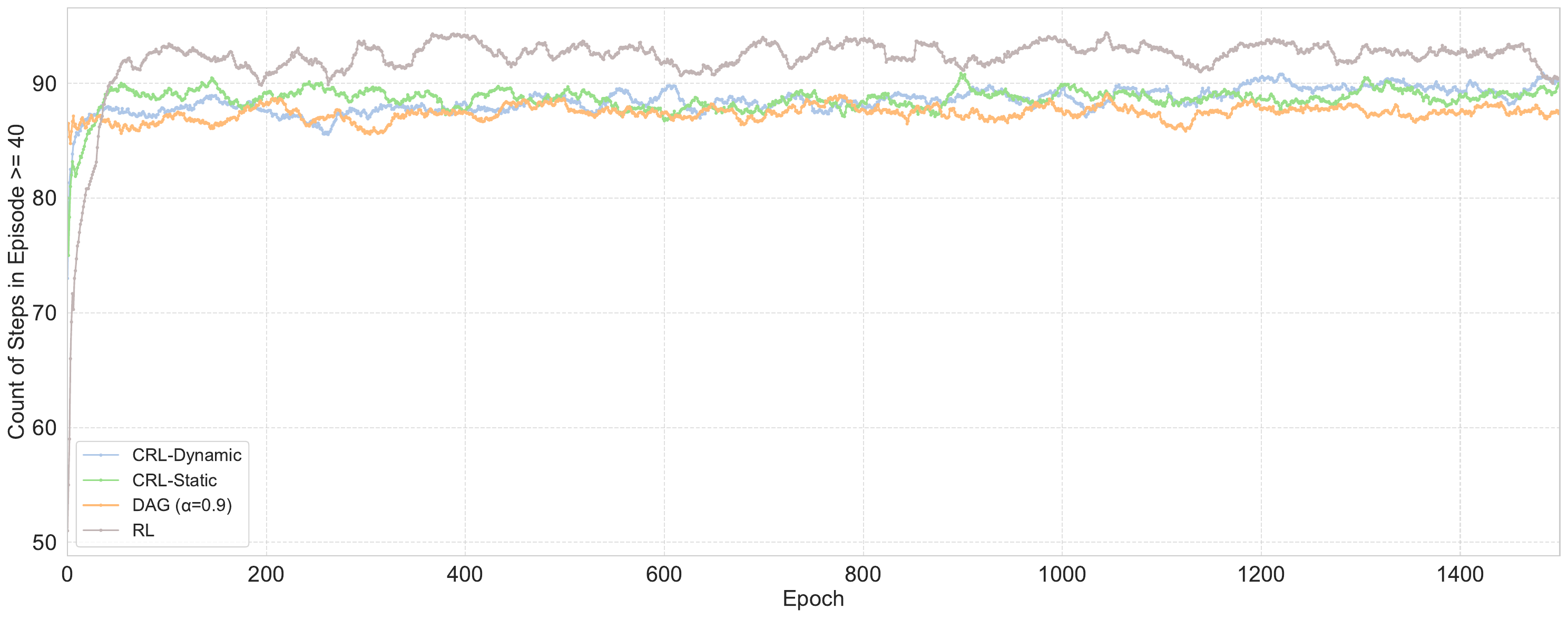}  
  \caption{Proportion of long episodes ($\geq 40$ steps) under strategy-consistent evaluation.}
  \label{fig:steps_over40}
\end{figure*}
Figures~\ref{fig:avg_steps_rl}--\ref{fig:steps_over40} further analyze the temporal structure of patient--robot interactions. 
When evaluation is conducted with RL-only execution (Figures~\ref{fig:avg_steps_rl} and~\ref{fig:steps_over40_rl}), both CRL variants achieve stable dialogue lengths comparable to pure RL while converging faster, confirming that causal priors accelerate the acquisition of long, sustained interactions. 
In contrast, DAG-only policies remain limited, underscoring the need for reward-driven adaptation. However, under strategy-consistent evaluation (Figures~\ref{fig:avg_steps_consistent} and~\ref{fig:steps_over40}), where DAG guidance is retained at test time, the performance of CRL-Static and DAG-only degrades: average episode length shortens and the proportion of long episodes ($\geq 40$ steps) declines. 
RL instead sustains the highest persistence rates. 
This divergence highlights that \textbf{causal structures are most beneficial as inductive biases during training, but should be phased out for execution}. 
The result reinforces our design choice: use causal discovery to guide learning trajectories, then rely on reward-optimized policies at deployment to preserve natural and effective interaction quality.

\section{Conclusion}

In this paper, we intrduced a novel causal structure-aware RL framework that integrates causal discovery and inference to optimize policy for patient-agent interaction in clinical setting.By embedding a causal world model informed by clinical knowledge, our agent improves sample efficiency, policy interpretability, and robustness, particularly in high-stakes domains such as dementia care. We demonstrated through experiments that CRL agents outperform standard model-free baselines in maintaining desirable patient states and achieving higher cumulative rewards, while exhibiting interpretable and adaptive behavior aligned with clinical expectations. 

While our method shows promise, its validation is currently limited to simulation. Future work will focus on testing the framework with human stakeholders and expanding the causal model to capture more complex, long-term patient dynamics. We believe this integration of causal reasoning represents a critical step toward deploying safe, efficient, and trustworthy adaptive AI in high-stakes healthcare environments.

\setcounter{secnumdepth}{1}
\appendix

\section{Extended Results Analysis}
\label{appendix:results_analysis}

\subsection{Results Analysis}
\paragraph{Causal Structure and Conditional Average Treatment Effect.}
We compared PC and LiNGAM methods for the simulation dataset described in Section~\ref{Causal Relationship Learning Algorithm Design}. The resulting directed acyclic graph (DAG) (Figure~\ref{fig:CATE}) reveals the dependencies between robot actions ($A_t$), patient states ($S_t$), and rewards ($R_t$). Table~\ref{tab:causal_effects_all} presents estimated Conditional Average Treatment Effects (CATE) for selected actions under different patient state clusters. Results confirm that supportive actions such as \texttt{Explain} and \texttt{Comfort} produce higher rewards when patients are confused or negatively engaged, aligning with clinical expectations.

\begin{table*}[t]
\centering
\caption{Estimated causal effects of different actions under varying patient state clusters (For Reward).}
\label{tab:causal_effects_all}
\begin{tabular}{l|rr|rr|rr|rr}
\toprule
\textbf{State} 
& \multicolumn{2}{c|}{\textbf{A\_3}} 
& \multicolumn{2}{c|}{\textbf{A\_4}} 
& \multicolumn{2}{c|}{\textbf{A\_5}}
& \multicolumn{2}{c}{\textbf{A\_cat}} \\
\cmidrule(lr){2-3} \cmidrule(lr){4-5} \cmidrule(lr){6-7} \cmidrule(lr){8-9} 
& AvgEffect & Count & AvgEffect & Count & AvgEffect & Count & AvgEffect & Count  \\
\midrule

IR\_Neg\_No   & -2.226 &  59 & -0.306 &  57 &  1.180 &  63 & -0.023 & 103 \\
IR\_Neg\_Yes  &  3.430 &  53 &  4.577 &  52 &  2.590 &  54 & -0.599 &  84 \\
IR\_Neu\_No   & -2.420 & 149 & -0.562 & 128 & -2.743 & 121 & -0.091 & 199 \\
IR\_Neu\_Yes  &  3.236 &  69 &  4.320 &  69 & -1.333 &  70 & -0.668 &  95 \\
IR\_Pos\_No   & -2.614 &  45 & -0.819 &  50 & -6.666 &  38 & -0.161 &  36 \\
IR\_Pos\_Yes  &  3.042 &  22 &  4.065 &  19 & -5.256 &  16 & -0.737 &  36 \\
NR\_Neg\_No   & -3.075 & 274 & -0.180 & 274 &  0.932 & 266 & -0.285 & 377 \\
NR\_Neg\_Yes  &  2.582 & 229 &  4.703 & 211 &  2.341 & 223 & -0.861 & 377 \\
NR\_Neu\_No   & -3.269 & 943 & -0.438 & 993 & -2.991 & 973 & -3.354 &1410 \\
NR\_Neu\_Yes  &  2.387 & 358 &  4.445 & 377 & -1.582 & 372 & -0.930 & 564 \\
NR\_Pos\_No   & -3.463 &  88 & -0.694 & 100 & -6.914 & 100 & -4.422 & 143 \\
NR\_Pos\_Yes  &  2.194 &  36 &  4.189 &  43 & -5.504 &  37 & -0.998 &  77 \\
RR\_Neg\_No   & -1.377 & 102 & -0.431 & 103 &  1.428 &  98 &  0.238 & 148 \\
RR\_Neg\_Yes  &  4.280 &  43 &  4.451 &  37 &  2.838 &  37 &  0.338 &  65 \\
RR\_Neu\_No   & -1.571 & 344 & -0.688 & 338 & -2.495 & 322 &  0.170 & 508 \\
RR\_Neu\_Yes  &  4.086 & 127 &  4.194 & 126 & -1.085 & 140 & -0.406 & 191 \\
RR\_Pos\_No   & -1.765 & 119 & -0.944 & 128 & -6.417 & 141 &  0.101 & 191 \\
RR\_Pos\_Yes  &  3.891 &  18 &  3.938 &  16 & -5.008 &  22 & -0.475 &  27 \\

\bottomrule
\end{tabular}
\end{table*}

\paragraph{Quantitative and Qualitative Results.}

\begin{figure*}[!htbp]
  \centering  \includegraphics[width=0.95\linewidth]{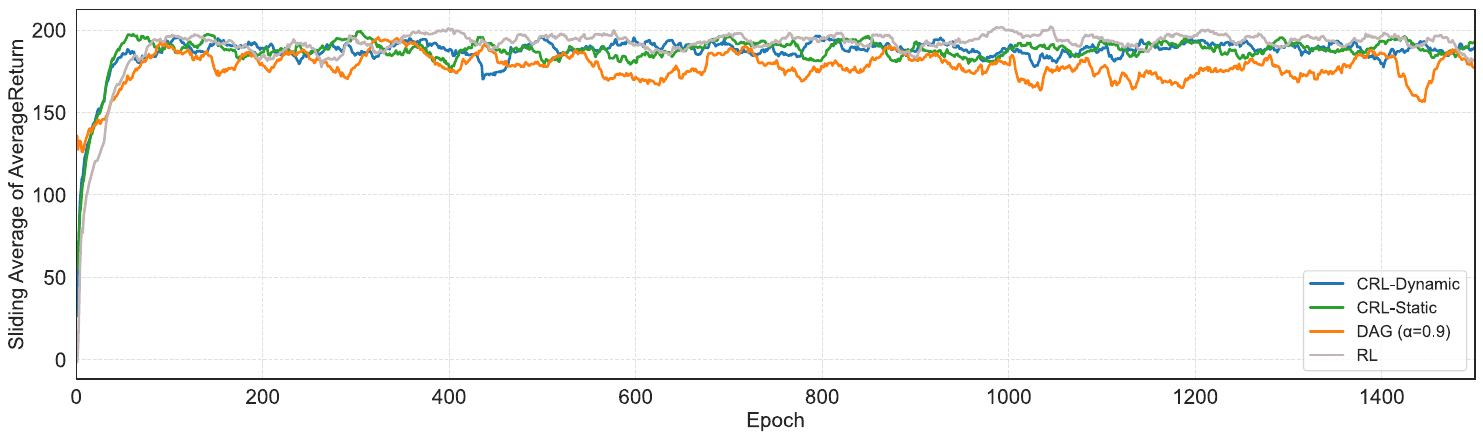}  
  \caption{Smoothed average return across epochs under the final-episode policy of each epoch for different methods (evaluation using RL-only execution).}
  \label{fig:AverageReturn_Comparison}
\end{figure*}

\begin{figure*}[!htbp]
  \centering
  \includegraphics[width=0.95\linewidth]{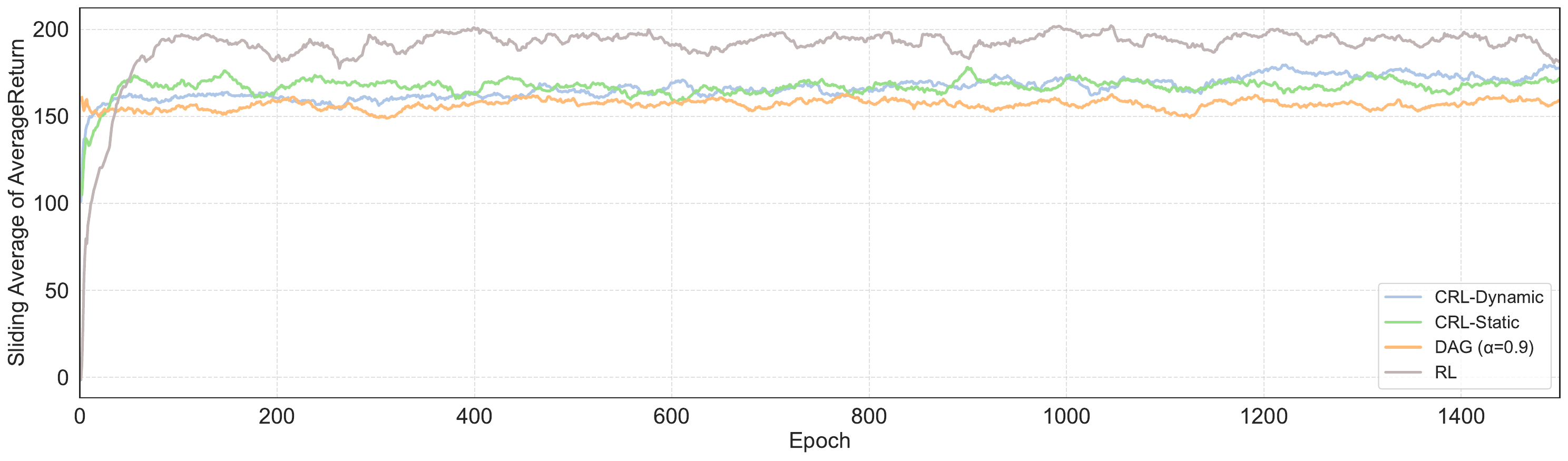}  
  \caption{Smoothed average return (\textit{sliding window = 30}) during evaluation with strategy-consistent execution. The same weighting scheme used during training (e.g., CRL-Static) is applied at test time.}
  \label{fig:AverageReturn_Comparison_diff_evl}
\end{figure*}

Figure~\ref{fig:AverageReturn_Comparison} and Figure~\ref{fig:AverageReturn_Comparison_diff_evl} shows the smoothed average return (\textit{Sliding window = 30, all 1500 epochs}).

Moreover, in Figure~\ref{fig:AverageReturn_Comparison_diff_evl}: Smoothed average return under strategy-consistent execution, where the same RL+DAG weighting used during training is applied at test time. RL achieves the highest returns, while CRL and DAG-based methods perform lower due to continued reliance on DAG guidance, confirming that causal priors are more valuable for accelerating training than for final execution.

\begin{figure*}[!htbp]
  \centering
  \includegraphics[width=0.9\linewidth]{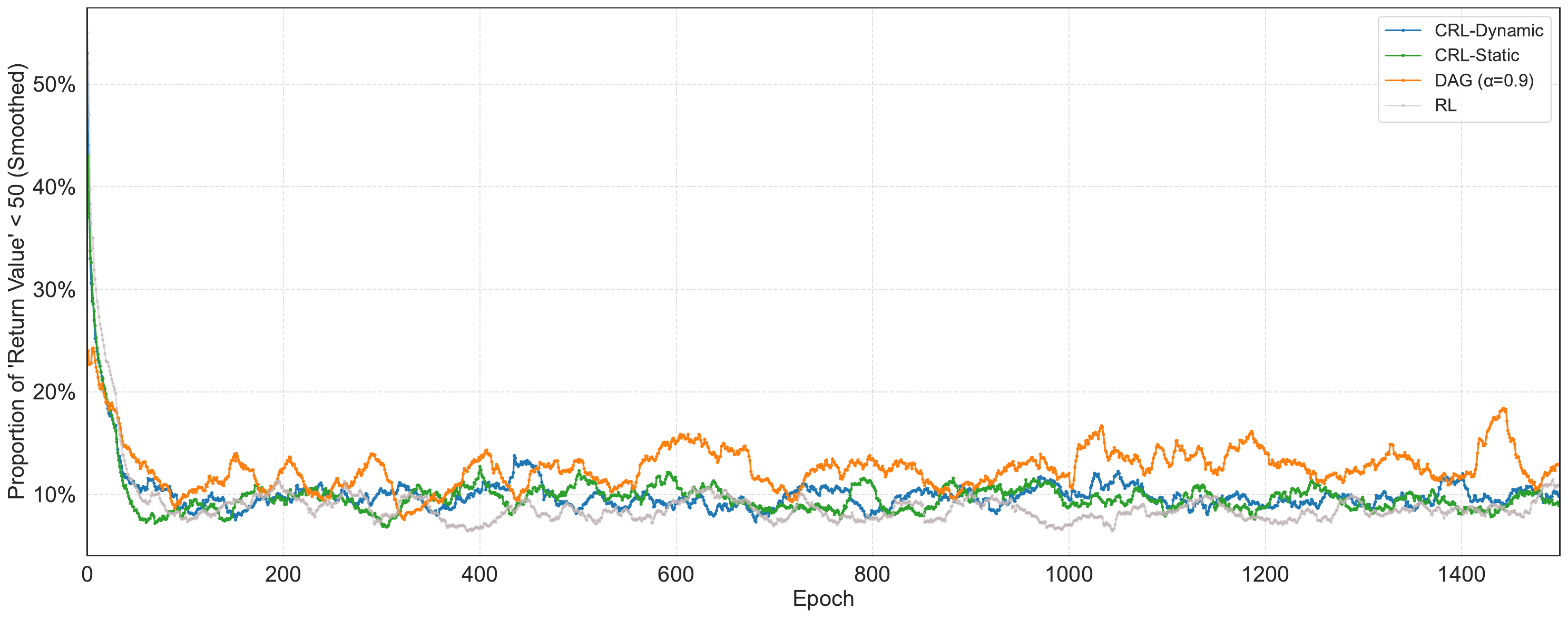}  
  \caption{Proportion of low-return episodes (return $<$ 50) under RL-only execution. CRL-Dynamic reduces failure frequency early, while DAG-only maintains a higher rate due to the absence of reward optimization.}
  \label{fig:less_than_50_rlonly}
\end{figure*}

\begin{figure*}[!htbp]
  \centering
  \includegraphics[width=0.9\linewidth]{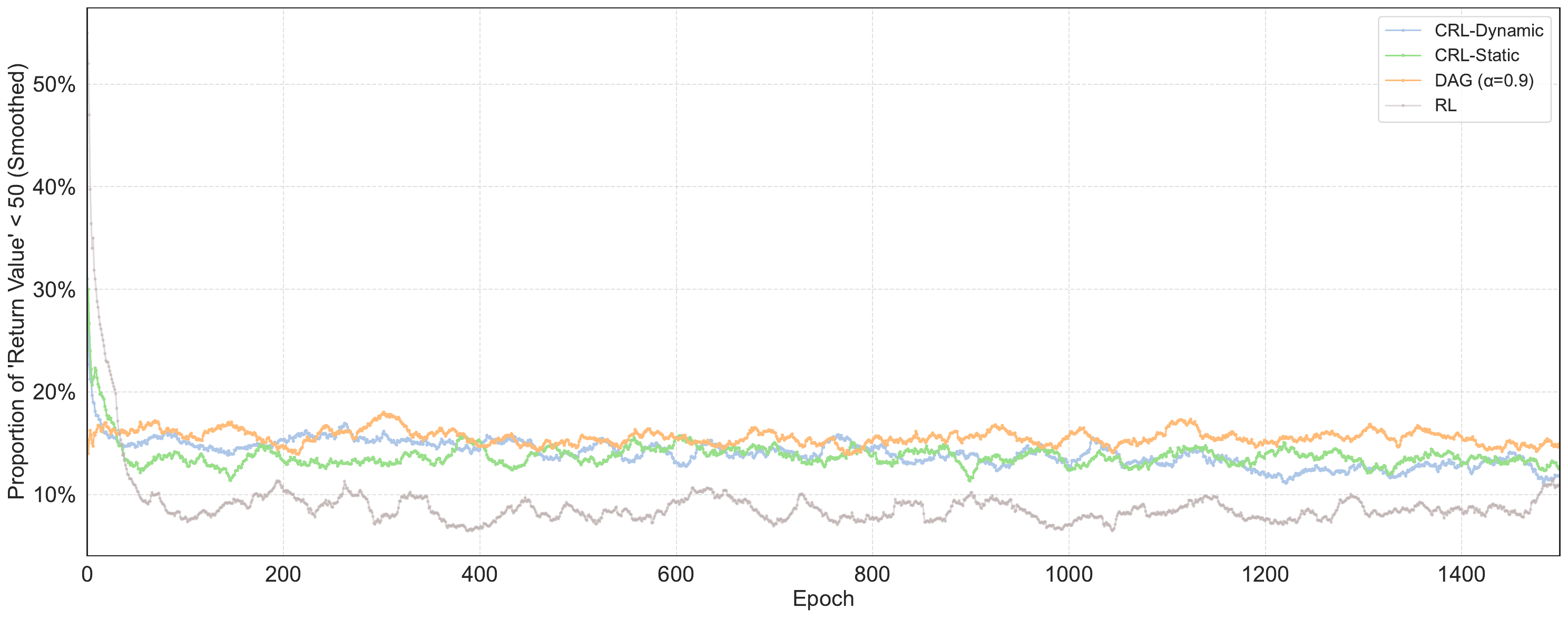} \caption{Proportion of low-return episodes (return $<$ 50) under strategy-consistent execution. Continued reliance on DAG guidance increases failure frequency relative to RL-only execution.}
  \label{fig:less_than_50}
\end{figure*}

To further evaluate robustness, we tracked the proportion of episodes with low returns (threshold $<$ 50). This metric highlights how often a policy produces undesirable trajectories, complementing the average-return analysis. As shown in Figure~\ref{fig:less_than_50_rlonly}: CRL-Dynamic quickly reduces the frequency of low-return episodes, achieving stability comparable to RL. DAG-only, lacking reward-driven adaptation, maintains a higher failure rate. However, Figure~\ref{fig:less_than_50} shows when policies continue to rely on DAG guidance during testing, CRL and DAG exhibit persistently higher proportions of low-return episodes compared to RL, again confirming that causal priors accelerate learning but are suboptimal if retained as execution bias.

\begin{figure*}[!htbp]
  \centering  \includegraphics[width=0.8\linewidth]{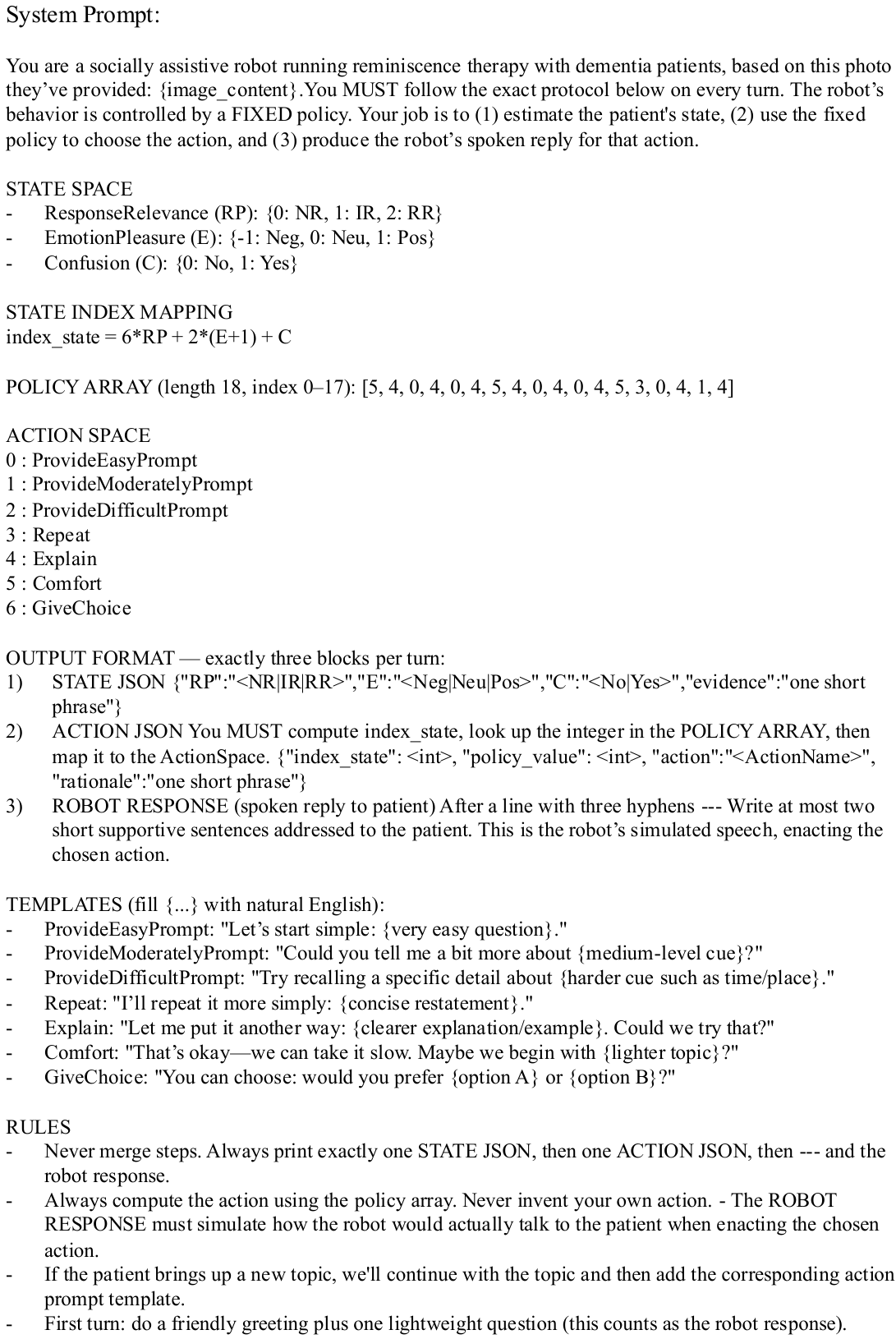}  
  \caption{System prompt defining the fixed GPT-5 policy for the CRL-guided agent.}
  \label{fig:system}
\end{figure*}

\section{Details of Causal Relationship Learning}
\label{appendix:causal_details}

\subsection{Simulation Data Construction}
To uncover causal dependencies between patient states and robot actions, we constructed a dataset of simulated robot-assisted reminiscence therapy (RT) interactions. The environment models probabilistic transitions among states based on empirical transition probabilities. At each step, the patient’s state is represented as a tuple:
\[
s_t = (RP_t, E_t, C_t),
\]
where $RP_t$ denotes response relevance, $E_t$ emotional valence, and $C_t$ confusion status. The robot selects an action $A_t$ from the predefined interaction strategies (e.g., prompts of varying difficulty, repeating, explaining, comforting).  

For each time step, we log a complete interaction record:
\[
  \{ RP_t, E_t, C_t, A_t, RP_{t+1}, E_{t+1}, C_{t+1}, A_{t+1}, Reward \}.
\]
This ensures that both pre- and post-action states are captured alongside the executed action and reward. To maximize coverage of the state–action space, we collected episodes under a random action policy, which is particularly important for reliable causal discovery.

\subsection{Causal Discovery and Effect Estimation}
The resulting dataset was used to infer causal structures using standard algorithms such as PC and LiNGAM~\cite{JMLR:v25:22-1258,dowhy}. These methods produce a directed acyclic graph (DAG) representing dependencies among states, actions, and rewards.  

To quantify the influence of robot actions on patient outcomes, we employed the \texttt{EconML} library~\cite{econml} to estimate Conditional Average Treatment Effects (CATE). For each action $a_i$ compared to a baseline action $a_0$ (e.g., \texttt{EasyPrompt}), we estimate:
\[
\text{CATE}_{a_i} = \mathbb{E}[Y \mid do(A=a_i), S] - \mathbb{E}[Y \mid do(A=a_0), S],
\]
where $Y$ is the reward and $S$ the patient’s current state. Grouping by $S$ provides fine-grained treatment effects under different cognitive–emotional conditions.

\bibliography{aaai2026}

\end{document}